\documentclass[10pt,twoside,twocolumn,a4paper]{article}

\usepackage[accepted]{bpasts}

\usepackage{t1enc}
\usepackage[utf8]{inputenc}
\usepackage{amsmath,amsfonts}

\usepackage{graphicx}
\usepackage{flushend}
\usepackage{cite}
\usepackage{graphicx}
\usepackage{mathtools}
\usepackage{amsmath}
\usepackage[utf8]{inputenc}
\usepackage{setspace}
\usepackage{enumitem}
\usepackage{hyperref}
\usepackage{graphicx}
\usepackage[toc,page]{appendix}
\usepackage{wrapfig}
\usepackage[justification=centering]{caption}
\usepackage{color}
\usepackage{multirow}
\usepackage{soul,xcolor}
\usepackage{setspace}

\abtitle{3D Face Reconstruction with Region Based Best Fit Blending}
\title{3D Face Reconstruction with Region Based Best Fit Blending Using Mobile Phone for Virtual Reality Based Social Media}

\abauthor{Anbarjafari et al.}
\author{Gholamreza ANBARJAFARI $^{1,2}$, Rain Eric HAAMER $^{1}$, Iiris L\"USI $^{1}$, Toomas TIKK $^{1}$, and Lembit VALGMA $^{1}$\email{shb@icv.tuit.ut.ee}}

\Address{$^1$ iCV Research Group, Institute of Technology, University of Tartu, Tartu 50411, Estonia\\
$^2$ Department of Electrical and Electronic Engineering, Hasan Kalyoncu University, Gaziantep, Turkey }

\Abstract{The use of virtual reality (VR) is exponentially increasing and due to that many researchers has started to work on developing new VR based social media. For this purpose it is important to have an avatar of the users which look like them to be easily generated by the devices which are accessible, such as mobile phone. In this paper, we propose a novel method of recreating a 3D human face model captured with a phone camera image or video data. The method focuses more on model shape than texture in order to make the face recognizable. We detect $68$ facial feature points and use them to separate a face into four regions. For each area the best fitting models are found and are further morphed combined to find the best fitting models for each area. These are then combined and further morphed in order to restore the original facial proportions. We also present a method of texturing the resulting model, where the aforementioned feature points are used to generate a texture for the resulting model.%\footnote{codes can be downloaded at: https://github.com/DeadScarab/FaceScanCodes}.
}

\Keywords{facial modeling, morphing, stretching, computer vision, similarity calculation, deformable model, 3D faces, virtual reality, blendshape}

\vol{XX} \no{Y} \year{2017}
\doi{10.1515/bpasts-2017-00ZZ}

\begin{document}
\maketitle

%%%%%%%%%%%%%%%%%%%%%%%%%%%%%%%%%%%%%%%%%%%%%%%%%%%%%%%
%% BODY OF THE ARTICLE

\section{Introduction}
\label{intro}

\noindent Reconstruction 3D face models has been a challenging task for the last two decades, because even a small amount of changes can have a huge effect on the recognizability  and accuracy of the generated model. Therefore, perfectly modelling a 3D human face still remains one of the quintessential problems of computer vision.  Also with the rapid increase of applications of virtual reality (VR) in gaming as well use use of VR application for making VR based social media, the task of making real-time 3D model of head which can be later mounted on avatar bodies have become very important \cite{olson2011design,santos2009head,kim2007virtual,anbarjafari2015objective,fernandez2017access,zeng2010social,trenholme2008computer}.
So far the easiest, fastest and most accurate methods have benefited from depth information that can be captured with recording devices like RGB-D cameras \cite{avots2016automatic,yamasaki2015fast,ding2016rgbd,huang2016human,valgma2016iterative,ding2016survey,lusi2017mimicking} or 3D scanners \cite{blanz2007fitting,traumann2015accurate,daneshmand2016real,fateeva2014applying}. However, nowadays development has been directed towards mobile devices, which limited to only using RGB information.

As an alternative commodity mobile phones that have their own on-device stereo cameras~\cite{kolev2014turning,ondruvska2015mobilefusion,tanskanen2013live,zhu2017role} can be used to recreate the depth data. Models created with passive multi-view stereo~\cite{maninchedda2016face,jain2016improving} have distinct features but very rough surfaces. Due to the noisiness of the input data heavy smoothing needs to be applied, resulting models that don't have finer details around more complex features like the eyes and nostrils.

%Existing 3D face model reconstructing algorithms that only use image or video data rely on a wide variety of different methods.
For 3D face reconstruction based on regular input images or videos, a large variety of methods and algorithms have been developed.
The most conventional approach has been to use a 3D morphable head model~\cite{blanz1999morphable,wood20163d}. Even though each researcher has used different metrics and methods, the main idea has been to minimize the error between input image and the 2D render of the model. The models features are iteratively morphed to minimize the difference between the input image and the image with the render of the model, using suitable lighting and orientation,  overlaid on the original face.  A huge downside to these methods is that they require a lot of processing power and time.~\cite{blanz1999morphable,blanz2003face}.
Another popular method is silhouette extraction, where the outer silhouette of a face is detected in video frames and the base model(without texture) is then constructed iteratively~\cite{lewis2014practice, baumberger20143d}. In~\cite{dou2014robust} and~\cite{choi20103d} landmark based shape reconstruction of 3D faces, which is very similar to RGB-D reconstruction, is used. In these cases features extracted using SIFT, SURF etc. were utilized. In some cases, the missing depth info was approximated using shading and lighting source estimation~\cite{kemelmacher20113d}. All of the above mentioned methods suffer from high complexity, which can result in longer computational time.

More efficient methods for face reconstruction generally rely on the detection of facial feature points or sparse landmarks and then stretching of a generic base model to produce a realistic output~\cite{zhang20123d,lin2010three,fan20103d,wu2016fast}. Among those are some that have even been specifically designed for mobile applications, but unfortunately the produced models tend to lack distinct features and the end result looks generic~\cite{qu2014fast}. Even though recognizable result can be achieved using excellent texturing techniques, the base model itself looks nothing like the real person. The algorithm behind Landmark-based Model-free 3D Face Shape Reconstruction~\cite{van2013landmark} tries to address the problem of generic looking outputs by avoiding statistical models as the base structure. They managed to produce genuine-looking models, but as the output is untextured and not a simple quadrilateral mesh structure, it is unusable for consumer applications.

In this paper we present a computationally light method for creating a 3D face model from video or image input, by constructing models for 4 key regions and later blending them. The method benefits from a high quality 3D scanned model database consisting of 200+ male head. As a pre-processing step all 4 key areas are individually rendered for each entry. As a first step of the method, the detected face is separated from the frame and 68 feature points are extracted. The face is divided into the 4 key regions, which are matched up against corresponding rendered areas of the database. These comparisons yield weights that are then used to combine a model from the predefined face regions. A stretching method similar to the ones described in \cite{zhang20123d,lin2010three,fan20103d,wu2016fast} is the applied. The model is then further morphed using the 68 feature points and their corresponding vertexes on the model in order to restore facial proportions. 

A texture is created by aligning the input images using piece-wise affine transformations to fit the UV map of the model. As an output our proposed method creates a 3D model with accurate facial features and fixed vertexes. This enables rigging, simple integration into existing systems or reintegration into the database while using minimal computational power. As a prerequisite, our method places some constraints on the input images and video, namely full visibility of the face with minimal facial hair and somewhat fixed lighting conditions.

\section{Proposed method}

\begin{figure*}[ht]
    \centering
    \includegraphics[scale = 0.75]{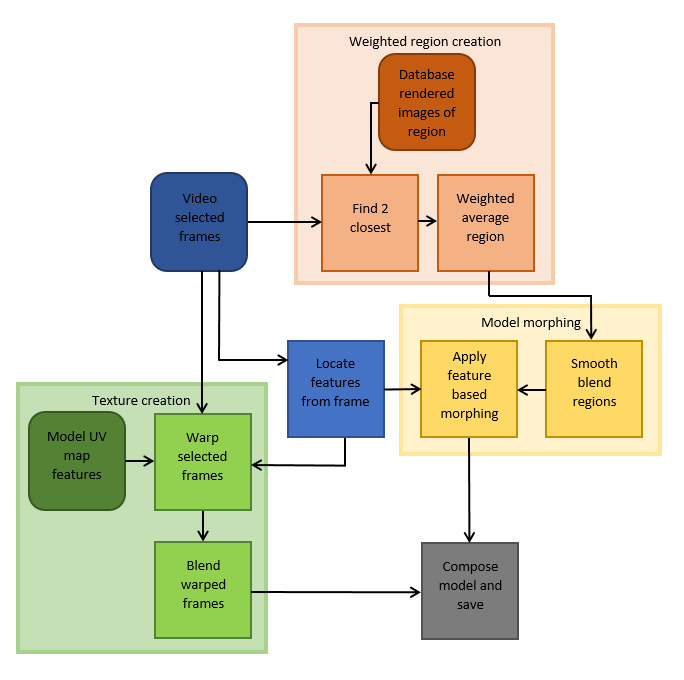}
    \caption{Block diagram describing the overall process}
    \label{fig:overall}
\end{figure*}

\noindent In this section a novel methodology for making 3D models from videos or image sequences is proposed. This approach benefits from a unique reference database of 3D head scans.  As illustrated in Fig. \ref{fig:overall}, the process can be divided into four main parts: 
\begin{itemize}
    \item feature extraction
    \item selection closest models
    \item model creation
    \item texture creation.
\end{itemize}

\subsection{Database}
\subsubsection{Data description}
\noindent The preliminary database consisted of textured 3D scans of 217 mostly Caucasian male head area, that were morphed and simplified so all shared the same vertex mapping and count. All of the textures were also warped to fit the same UV map. 
In this step everything behind the head from the original scans was discarded, including the hair and the ears. In the original data some of the scans contained facial hair or severe noise/deformations. However the rest of the regions were usually not affected. 

The pre-processed model consists of a mesh that contains 6000 quadrilateral faces and a corresponding 2048x2048 RGB texture. The indexes of the vertexes in all of the models are ordered and don't vary from model to model.

\subsubsection{Rendering regions}
\begin{figure}
    \centering
    \includegraphics[scale = 0.19]{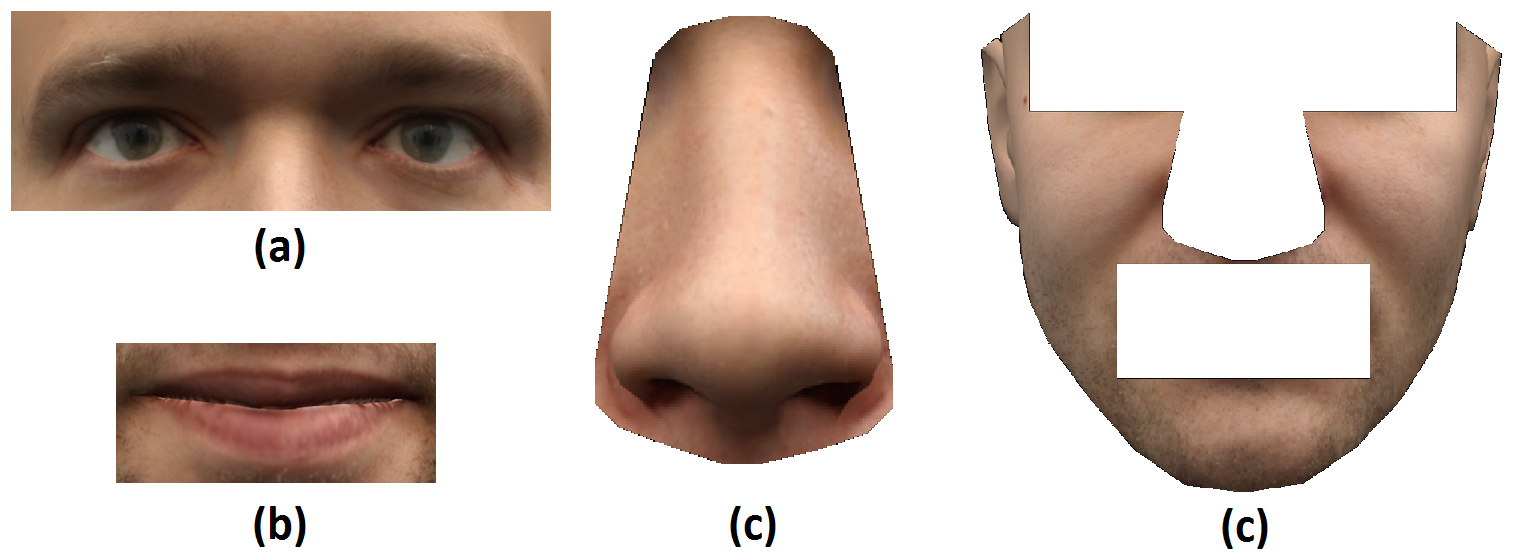}
    \caption{Cut out sections of a rendered model from the model database. The 4 sections represent the eyes (a), mouth (b), nose (c) and the rest of the face (d) regions.}
    \label{fig:render_regions}
\end{figure}

\noindent An auxiliary database of rendered face regions was made in order to find the weights of each model based on the input image. All of the models were rendered in frontal orientation using perspective cameras with 5 directional light sources from the front, sides and top - all aimed towards the center of the head. The rendered images were then cut into 4 sections which correspond to the eyes, nose, mouth regions and rest of the face as shown in Fig. \ref{fig:render_regions}.

\subsection{Feature extraction}
\begin{figure}
    \centering
    \includegraphics[scale = 0.4]{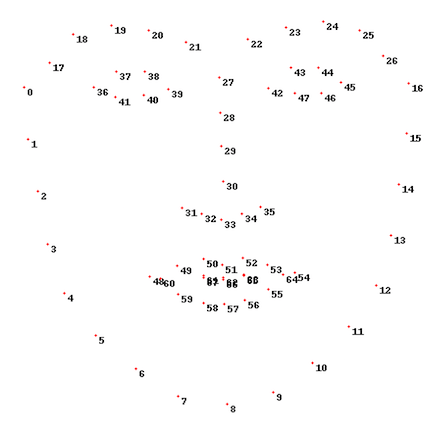}
    \caption{The 68 feature points that are extracted}
    \label{fig:open}
\end{figure}

\noindent For the facial feature extraction we use facial analysis toolkit~\cite{Baltrusaitis2016} to estimate and extract $68$ facial feature points from the input video sequence. Due to limitations of the toolkit, the input face must have a near neutral face, with no eye-wear or thick facial hair which obstruct the feature detection. These points are illustrated in \ref{fig:open}. The algorithm also outputs the rotation parameters, so this can be used to pick the frontally best aligned frame as the main reference for creating the model. This also allows alignment of the face to a frontal one in case of slight tilt or rotation.

As the next step the area containing feature points is extracted and the feature points from that area are normalized to fit between $0$ and $1$. 

\subsection{Region-based selection and weight calculation}
\noindent In this step we calculate similarities between database regions and the input grayscale regions. There are altogether three different similarity indexes used: PCA based measure, SSIM~\cite{wang2004image} (structural similarity index) and a LBP(local binary pattern) histograms  difference measure~\cite{ojala2002multiresolution}. 

In the PCA based approach the database images are first vectorized and the covariance matrix of these vectors is found. The principal components and score vectors are stored for future use. When looking for the closest match the corresponding image region is also vectorized and its score vector on the principal components is found. We treat the Eucleidean difference between the score vectors as the error and its' inverse as the similarity measure.

Since the SSIM finds the similarity between two images, the maximum being one, we use the dissimilarity ($1-ssuim$) as the error. In case of LBP features we also apply PCA to the output feature and find the distances as described earlier.

For picking the closest mouth and nose regions, SSIM was used. For eyes we used LBP as they have the most high frequency data and for face shape we used PCA based approach as the placement zero and non-zero values has high impact on this measure.

For each region a similarity function between database rendered regions and the regions extracted from the frontal face is applied. After which the corresponding similarity function, an error vector is obtained for each region. Let us denote it by $E = (E_1,E_2,\ldots , E_n)$, where $n$ is the number of reference heads in the database. Let $I$ denote the set of indexes of the database models corresponding to smallest error. The weights $W_i, i\in I$ are calculated according to the formula:
\begin{equation}
W_i =  \frac{E_i^{-1}}{\sum_{j\in I}{E_j^{-1}}}
\end{equation}

The rest of the weights are set to zero: $W_i=0$, $ \forall i \not \in I $

Based on the errors of the similarity three (or in some cases 1) closest matches are picked for each region, and weights that are inversely proportional to the errors are assigned. The weights are normalized to they sum to 1.

\subsection{Blended model creation}

\begin{figure}
    \centering
    \includegraphics[scale = 0.2]{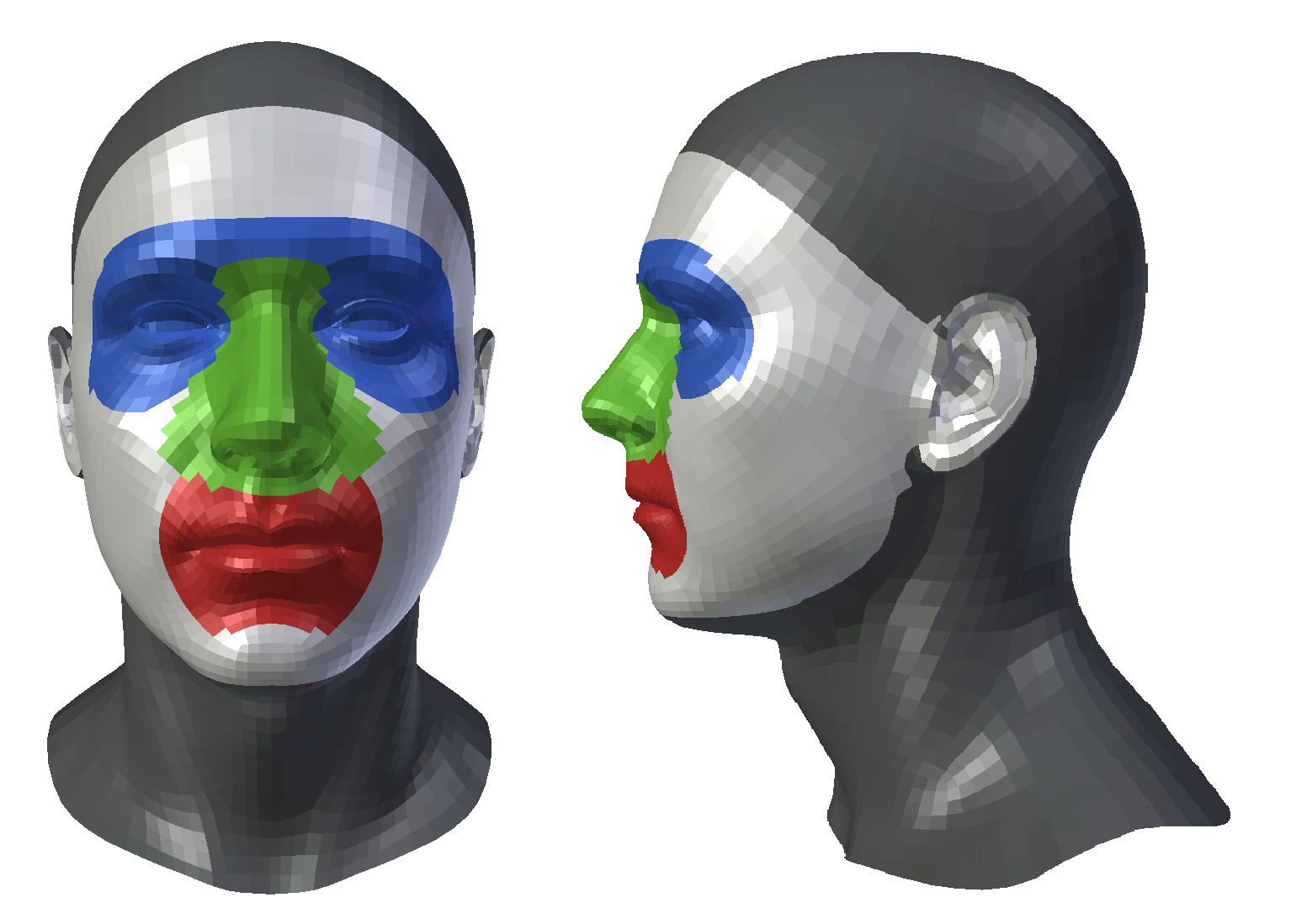}
    \caption{The 5 primary regions as shown in colors are used for linear weighted combinations to generate the base model. The eye region is shown in blue, the nose region is shown in green, the mouth region is shown in red, the rest of the face region is shown in white and the unused parts of the model are shown in gray.}
    \label{fig:model_regions}
\end{figure}

\noindent The model is separated into 5 primary regions as shown in Fig. \ref{fig:model_regions}, out of these regions, only the eyes, nose, mouth regions and rest of the face area are used. For each region 3D models $AM$ are created as blendshapes~\cite{lewis2014practice} from weighted combinations of an array of models $M = (M_0, M_1, M_2, \dots , M_n)$
and their corresponding weights $W = (W_0, W_1, W_2, \dots , W_n)$, as in formula \ref{AM}. 
Each model ($AM$) has its corresponding region of interest signified with an array of indexes called $I_A$.

Out of the four models created, the one representing the rest of the face will be called the base model $BM$. The $BM$ will be the starting model, on which the other weighted models are added one-by-one. 
The base model $BM$ also has its own corresponding index array called $I_B$. 

\begin{equation}
BM = M_i\cdot W_{i}^T 
\label{AM}
\end{equation}

\begin{figure}
    \centering
    \includegraphics[scale = 0.16]{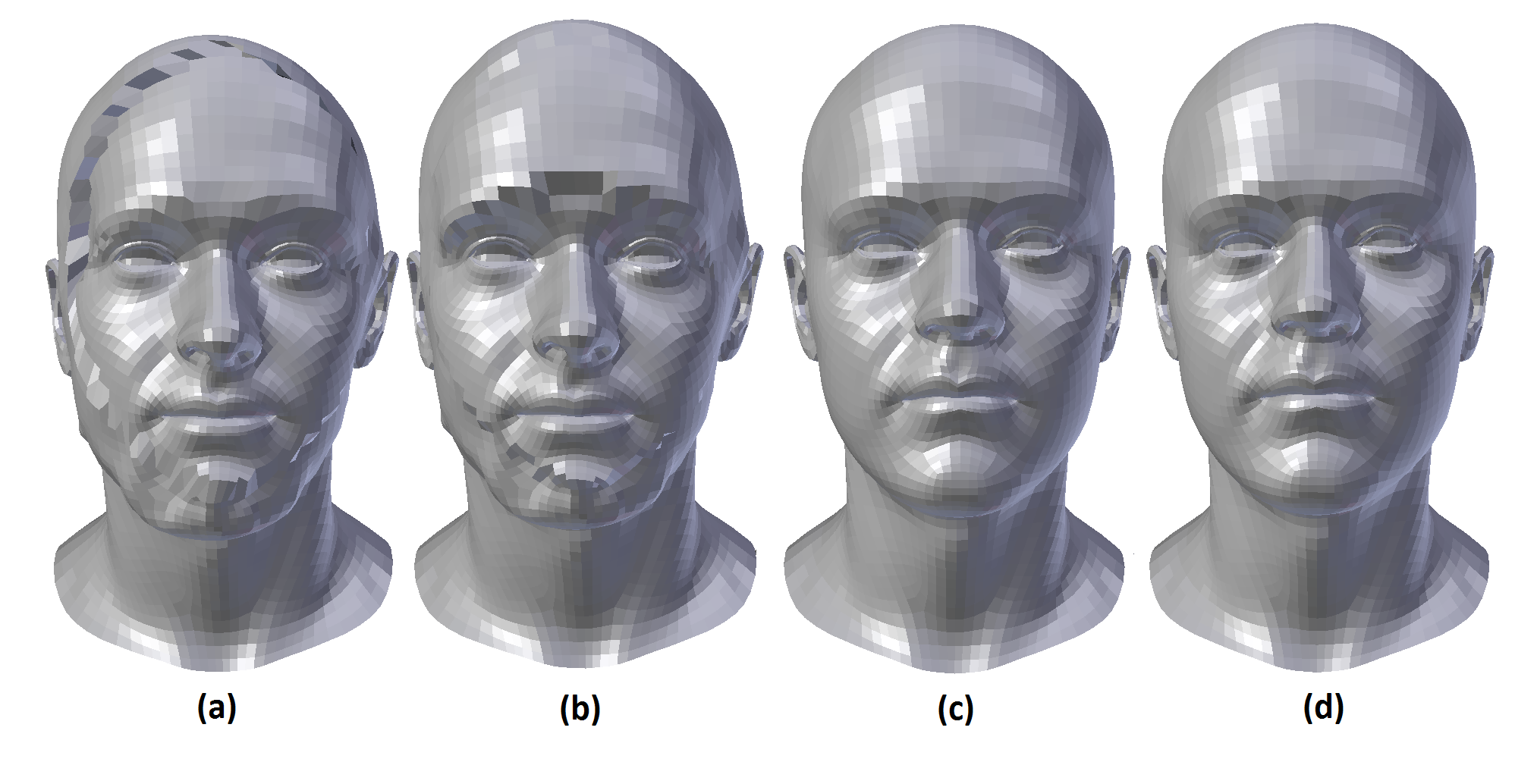}
    \caption{Processing steps for blending and morphing the 3D model regions to create an output model. The weighted model regions are combined to form the base shape of the output model (a). To ease the transitions between different regions the regions are first aligned with one another (b) and then the overlapping areas are blended together using a blending function (c). The final combined model is then further morphed (d) to fit the input image}
    \label{fig:process}
\end{figure}

Then regions from these composite 3D models are combined together to form the overall base shape of the 3D model as shown in Fig. \ref{fig:process}a.

To minimize the effect of large transitions between regions, mean locations are calculated for the overlapping vertexes of 2 combined regions. The vertex indexes for this new region will be denoted as $I_{AB} = I_A \cap I_B$.

The vector originating from the mean point of the added region to the mean of the existing region is found using formula \ref{v_BA}.

\begin{equation}
\vec s_{BA} = \frac{1}{|I_{AB}|} \cdot \sum_{i \in I_{AB}}(v_i^B-v_i^A)
\label{v_BA}
\end{equation}

To reduce the large effect of small locational variations that have later a big effect on the smoothing process, the vector $\frac{1}{2} s_{BA}$ is added to the points in the added region using the formula \ref{AM_j}.
\begin{equation}
v_i^A = v_i^A + \frac{\vec s_{BA} }{2} \quad i \in I
\label{AM_j}
\end{equation}

To further smooth the transition between different regions, a blending function is applied when calculating the model weights for each vertex. This can be seen as the transition from model (b) to (c) in Fig. \ref{fig:process}.
In order to scale the blending function for the added regions area, the maximum translations along the $X$ and $Y$ axes of the added region are calculated.
%In case of $v_i^A = (x_i^A,y_i^A)$

\begin{equation}
\Delta_x = \max_{i \in I_A}\{x_i^A\} - \min_{i \in I_A}\{x_i^A\}
\label{Deltax}
\end{equation}

Where $x_i^A$ is the $X$ axis component of $v_i^A$.
$\Delta_y$ is found using a similar function as \ref{Deltax} and they will be denoted as $\Delta = (\Delta_x, \Delta_y)$.

The distances of all of the added region points from the their mean can then be calculated and normalized using the function \ref{D_j}.From now on $j$ will denote the index of a vertex from the model.

\begin{equation}
\delta_j = || \frac{\mu_{AM_{I_A}} - v_i^A}{\Delta}||  \quad i \in I
\label{D_j}
\end{equation}

As all of the distances will be in reference to the mean of the added region, similar calculations will not be applied to the base region.

A transition weight function $\mathit{twf}(x)$ is then applied to each of the point distances to get the weights for each vertex. The two regions are then combined using:
\begin{equation}
\mathit{twf}(\delta_j) = 1.013 - \frac{1.019}{1+(\frac{\delta_i}{0.264})^{3.244}}
\end{equation}

\begin{equation}
BM_{j} = BM_{j}\cdot(1-\mathit{twf}(\delta_j)) + AM_{j}\cdot \mathit{twf}(\delta_j)
\end{equation}

Since the nose region has some overlapping areas with the eye and the mouth region, it is added to the base model first. Next the eye and mouth regions are added, though here the order plays no role as they don't share any vertexes.

\subsection{Model morphing}

\noindent The generated blended model $BM$ has features which resemble the desired face, but some of the proportions are off. Due to that the model is further morphed so key features match their actual locations.
The model has 68 vertexes mapped to 68 feature points $F = \{F_i\} \quad i \in \{1, 2, \dots , 68\}$ where value of $i$ will remain the same for all of the following functions.
For each mapped vertex a movement vector to the feature location is calculated as:

\begin{equation}
\vec f_{i} = F_i - BM_i
\end{equation}

For each of the points in the model, a distance is calculated from each mapped vertex and a adjustment weight function $\mathit{awf}(x)$ is applied to get the weight of each vector for each vertex.

\begin{equation}
d_{ji} = |F_i - BM_j|
\end{equation}

\begin{equation}
\mathit{awf}(d_{ji}) = 1 - \bigg({1+e^{\big(\frac{-d_{ji}}{\sigma_i^2\cdot k} + 0.5\big) \cdot 7}}\bigg)^{-1}
\label{awf}
\end{equation}

Where $\sigma_i$ defines the drop-off rate for the adjustment weight function, which is separately defined for each feature point, $k$ is a similar multiplier but it is independent from the feature point values.
The vectors $\vec f_{i}$ are then multiplied by the weights for each point and summed together:
\begin{equation}
\vec s_j = \sum_{i=1}^{n}(\mathit{awf}(d_{ji})\cdot \vec f_i)
\end{equation}

Then the sum of these weights $r_j$ is calculated by:
\begin{equation}
r_j = \sum_{i=1}^{n}\mathit{awf}(d_{ji})
\end{equation}
The resulting vector $s_j$ is further divided by the sum of the weights, if $r_j > 1$. This is done to keep densely packed features from overwhelming vertexes with their respective changes. The final vectors are added to every point in the model.

\begin{equation}
BM_j = 
    \begin{cases}
        BM_j + \vec s_j,                                              & \quad \mathrm {if}\quad r_j \leq 1 \\
        BM_j + \frac{\vec s_j}{ r_j }, & \quad \mathrm {if}\quad r_j > 1
    \end{cases}
\end{equation}

Since some feature vertexes define larger areas like the eyebrows and chin, while others define finer details like the eyes and nostrils, the features also have their own weights called $\sigma_i$. These weights allow for different features to have different adjustment weight functions as shown in the formula \ref{awf}.
The morphing is applied 3 times with 3 different $k$ values, starting with more general features and ending with smaller details, where the adjustment weight function sets the weights of points that have larger $\sigma_i$ values to zero. 

\subsection{Texture creation}
\begin{figure}
    \centering
    \includegraphics[scale = 0.03]{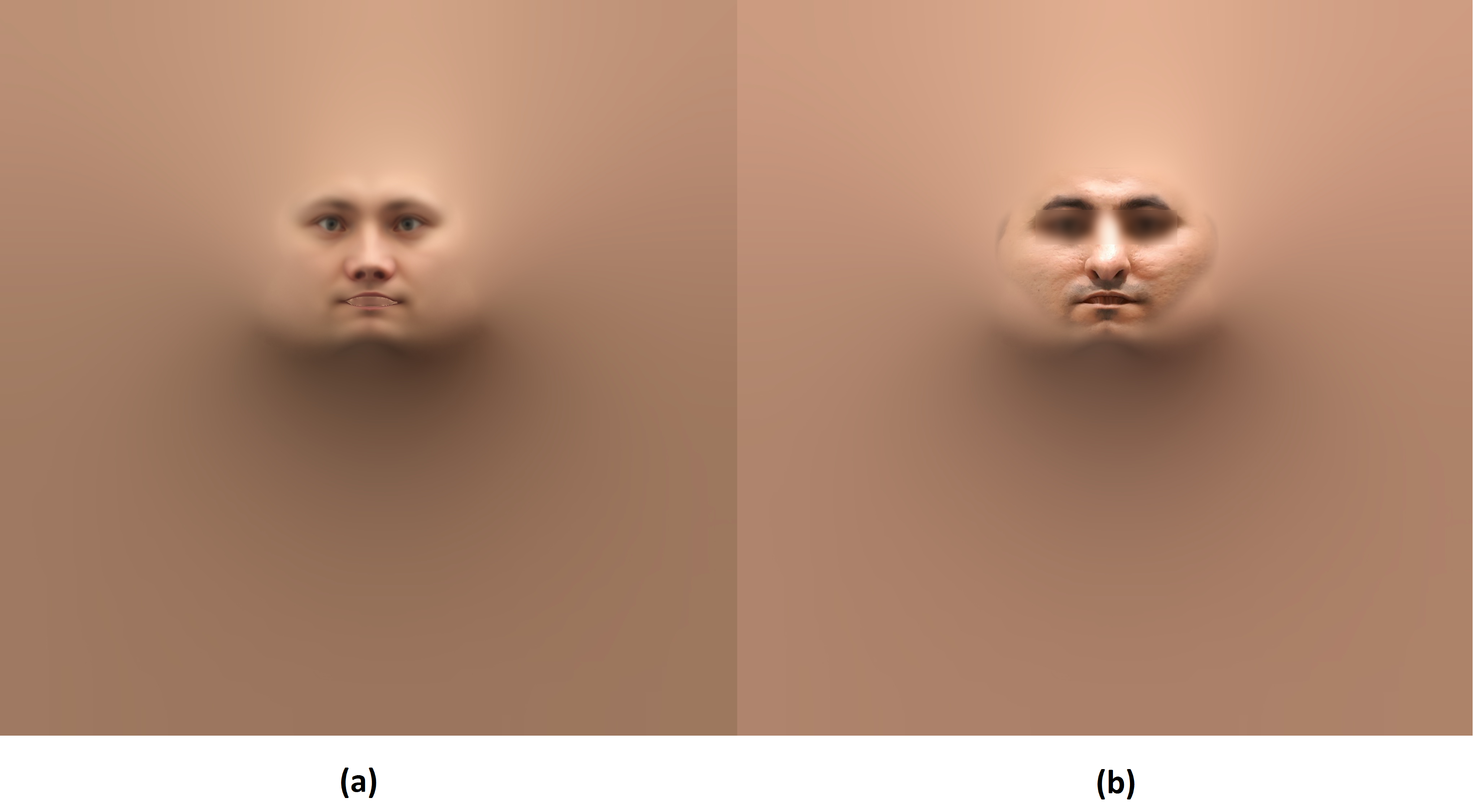}
    \caption{The average texture created from the existing textures in the database on the left and the input image fitted to the UV map on the left}
    \label{fig:texture}
\end{figure}

\noindent The texture of any 3D model is strongly bound by the UV map. By hand it is difficult to morph images into a form that would correspond to this map. 
In this work an average texture was first created based on the database textures Fig. \ref{fig:texture}a. Since all of the models in the database have a corresponding UV map, the input image had to be transformed to fit the same map. In order to warp the input image appropriately, $68$ vertexes on the UV map were marked corresponding to the $68$ feature points on the input image.

\begin{figure}
    \centering
    \includegraphics[scale = 0.1]{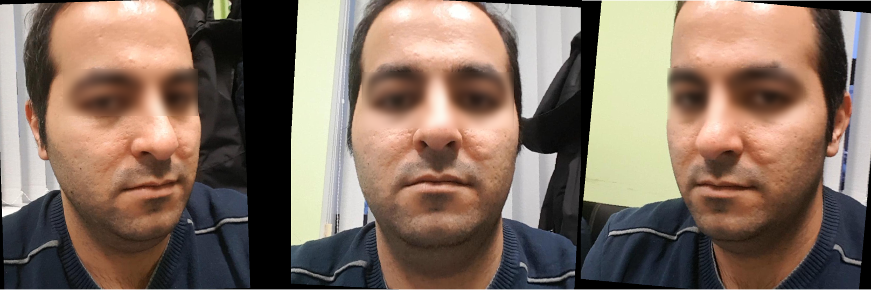}
    \caption{An example of images picked to combine}
    \label{fig:input_images}
\end{figure}

For texture creation we assume that there are no distinctive shadows present in the video sequence. Based on the extracted feature points and rotations, three frames are picked from the sequence based on yaw at approximately $30$ degrees on both sides and the frontal frame, as in figure\ref{fig:input_images}. From the face area also the median face color was extracted.

As the next step the average texture colors are shifted to match the median skin tone to assure a smooth and natural transition on the edges of the skin. Then the frames are piece-wise affinely transformed into an image of size $2048\times2048$ and blended together into a partial texture. This however has very sharp edges and lots of black pixels, so it is then blended only on the edges with the underlying and already shifted average texture. The average texture and the resulting texture can be seen in Fig. \ref{fig:texture}b.

\section{Experimental results}

\noindent We evaluated our method with several faces, some of which were included in the original scanned database. This was done in order to test the method's model selection and reconstruction proficiency.

\begin{figure}
    \centering
    \includegraphics[scale = 0.047]{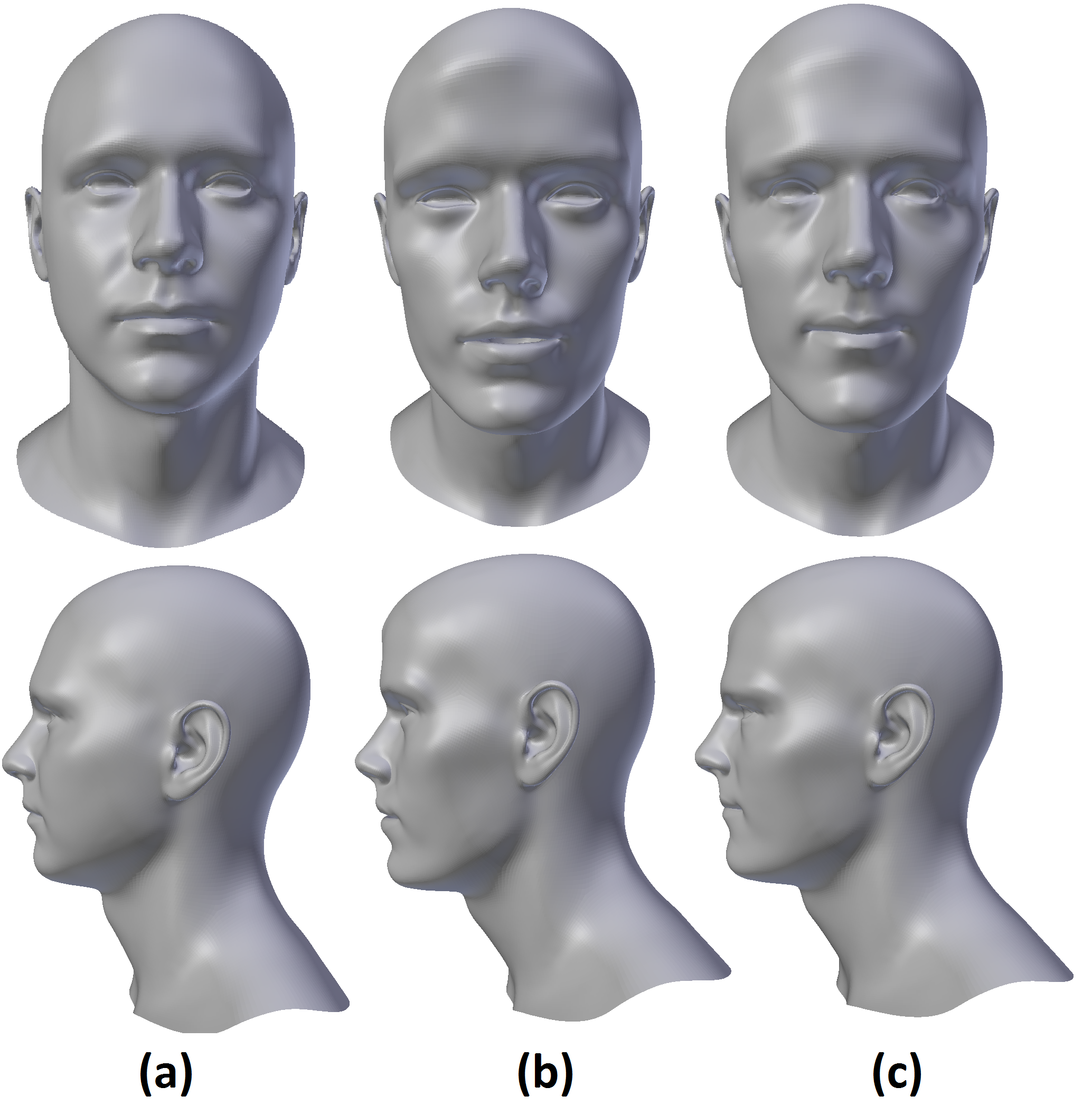}
    \caption{An average model from the scanned database (a), a scanned model (b) and the reconstructed version using a mobile phones camera (c).}
    \label{fig:compare}
\end{figure}

For the faces of people included in the 3D scanned database, the system selected most of the facial features that matched the person. In some cases a different mouth or a set of eyes were chosen when the scanned emotion and the emotion in the captured image were different. This is illustrated by Fig. \ref{fig:compare}, where the participant had a more surprised facial expressions during the scan and a neutral face during the recording. Regardless, the resulting models were nearly identical to the original scanned versions of those faces.

\begin{figure}
    \centering
    \includegraphics[scale = 0.095]{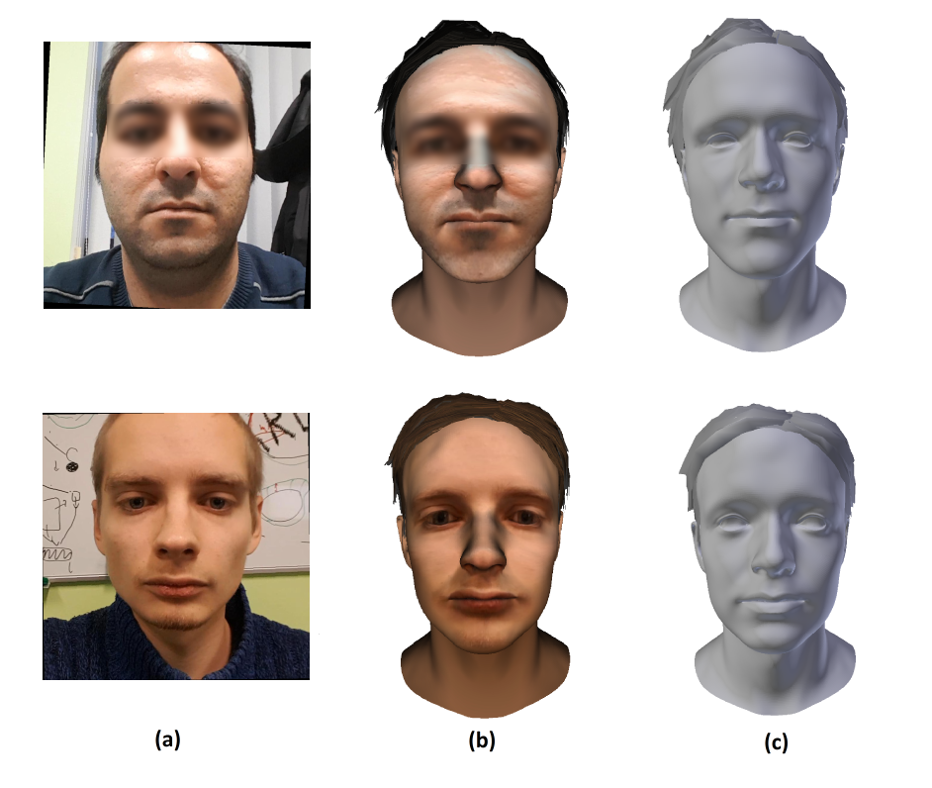}
    \caption{Textured (b) and untextured (c) models generated from the example input images (a) using the developed method. Both models have a manually created hair mesh. Neither participants had scanned counterparts in the model database.}
    \label{fig:models}
\end{figure}

In case of subjects that were not in the database, the system managed to choose facial features that were remarkably similar to the desired face. On occasions where no corresponding feature existed in the database, a blended version from 2 nearest features was created. The result of this blending were not always very similar to the original. This shortcoming can be seen  Fig. \ref{fig:models}, for the nose in the first row had no structurally similar nose in the entire database.

The goodness of the method was largely limited by the feature detection and the nature of the database. The database contained very little information on the general shape of the head and the feature detection method was very unreliable when it came to detecting the shape of larger areas like the chin. Because of that the resulting heads tended to have the same overall structure. This is very visible in case of the model in the first row in Fig. \ref{fig:models} that has a jawline which is smaller and sharper jawline than expected.

The recordings were conducted with a Samsung Galaxy S7 and the model creation process was run on an i7-4790@3.6GHz CPU with MATLAB 2015a. The whole process, including the recording of a 10 second clip, took about 3 minutes per model. For an i7@4.2GHZ with MATLAB 2016b the model generation took less than a minute.

\section{Conclusion}
\noindent In this paper, a method of recreating 3D facial models from portrait data,  which uses very little computational power in order to produce a recognizable facial model, was proposed. The main idea behind our system is finding the closest matching models for different facial regions and then combining them into a single coherent model. We have also presented a novel and simple method of texturing the resulting model by piece-wise affinely transforming input images to fit a desired UV map based on detected feature points. 

Our experimental evaluation has shown that our method is able to select the best corresponding facial features within a reasonable time frame. Our method has also partly overcome the problem of average looking models, which is a huge obstacle in methods that only use stretching to recreate shape. Unfortunately our method is still susceptible to problems relating to the overall shape of the head. This can be fixed with better facial feature extraction methods and with a database that includes a good selection of head shapes. The method can easily  be made more robust by rendering the regions with more refined parameters.

\section*{Acknowledgements}
This work has been partially supported by Estonian Research Council Grants (PUT638), The Scientific and Technological Research Council of Turkey (T\"UBİTAK) (Proje 1001 - 116E097) , the Estonian Centre of Excellence in IT (EXCITE) funded by the European Regional Development Fund and the European Network on Integrating Vision and Language (iV\&L Net) ICT COST Action IC1307.

\bibliographystyle{IEEEtran}

\bibliography{paper}

\end{document}